\documentclass[conference]{./IEEEtran}
\IEEEoverridecommandlockouts
\usepackage{cite}
\usepackage{amsmath,amssymb,amsfonts}
\usepackage{algorithmic}
\usepackage{graphicx}
\usepackage{textcomp}
\usepackage[svgnames]{xcolor}
\usepackage[colorlinks,breaklinks,urlcolor=black,linkcolor=black,citecolor=black]{hyperref}
\usepackage{todonotes}
\usepackage{cleveref}
\usepackage{listings}

\usepackage[acronym,toc,symbols]{glossaries}
\glsdisablehyper
\newacronym{iot}{IoT}{Internet of Things}
\newacronym{ros}{ROS}{Robotic Operating System}
\newacronym{k8s}{K8s}{Kubernetes}
\newacronym{lan}{LAN}{local area network}
\newacronym{cni}{CNI}{container network interface}
\newacronym{dds}{DDS}{Data Distribution Service}
\newacronym{rtps}{RTPS}{Real-Time Publish Subscribe}
\newacronym{qos}{QoS}{Quality of Service}
\newacronym{cncf}{CNCF}{Cloud Native Computing Foundation}

\newcommand\copyrighttext{%
  \footnotesize © 2024 IEEE.  Personal use of this material is permitted.  Permission from IEEE must be obtained for all other uses, in any current or future media, including reprinting/republishing this material for advertising or promotional purposes, creating new collective works, for resale or redistribution to servers or lists, or reuse of any copyrighted component of this work in other works.}
\newcommand\copyrightnotice{%
\begin{tikzpicture}[overlay, remember picture]
\node[anchor=south,yshift=10pt] at (current page.south) {\fbox{\parbox{\dimexpr\textwidth-\fboxsep-\fboxrule\relax}{\copyrighttext}}};
\end{tikzpicture}%
}

\def\BibTeX{{\rm B\kern-.05em{\sc i\kern-.025em b}\kern-.08em
    T\kern-.1667em\lower.7ex\hbox{E}\kern-.125emX}}
\begin{document}

\title{RoboKube: Establishing a New Foundation for the Cloud Native Evolution in Robotics
}

\author{\IEEEauthorblockN{1\textsuperscript{st} Yu Liu}
\IEEEauthorblockA{\textit{Ericsson Research} \\
Stockholm, Sweden \\
yu.a.liu@ericsson.com}
\and
\IEEEauthorblockN{2\textsuperscript{nd} Aitor Hernandez Herranz}
\IEEEauthorblockA{\textit{Ericsson Research} \\
Stockholm, Sweden \\
aitor.hernandez.herranz@ericsson.com}
\and
\IEEEauthorblockN{3\textsuperscript{rd} Roberto C. Sundin}
\IEEEauthorblockA{\textit{Ericsson Research} \\
Stockholm, Sweden \\
roberto.castro.sundin@ericsson.com}
}

\maketitle
\begin{abstract}
Cloud native technologies have been observed to expand into the realm of \gls{iot} and Cyber-physical Systems, of which an important application domain is robotics. In this paper, we review the cloudification practice in the robotics domain from both literature and industrial perspectives. We propose RoboKube, an adaptive framework that is based on the \gls{k8s} ecosystem to set up a common platform across the device-cloud continuum for the deployment of cloudified \gls{ros} powered applications, to facilitate the cloud native evolution in robotics. We examine the process of modernizing ROS applications using cloud-native technologies, focusing on both the platform and application perspectives. In addition, we address the challenges of networking setups for heterogeneous environments. This paper intends to serves as a guide for developers and researchers, offering insights into containerization strategies, ROS node distribution and clustering, and deployment options. To demonstrate the feasibility of our approach, we present a case study involving the cloudification of a teleoperation testbed.

\end{abstract}


\glsresetall
\section{Background}
\label{sec:back}
\copyrightnotice
Cloud native evolution
as defined by \gls{cncf}, is characterized by a shift from monolithic architectures to microservices, from manual deployment to continuous integration/continuous delivery (CI/CD), and from static infrastructure to dynamic, scalable, and resilient systems orchestrated by platforms like \gls{k8s} \cite{cncf}.
The success of cloud-native practices in the cloud industry has been brought into the IoT and edge computing domains, represented by promising projects like K3s, MicroK8s, KubeEdge, Azure IoT Edge, and Edgenesis Shifu, etc.

Meanwhile, \gls{ros}, specifically ROS 2, is the robotic community’s answer to the demand for a modular, scalable, and reliable architecture to build robotic applications such as sensing, planning, mobility, and autonomy. ROS 2 offers quality of service for communications, real-time support, and enhanced security features, all of which are critical for industrial applications. The development and widespread adoption of ROS have significantly accelerated innovation in robotics, reducing the barrier to entry and fostering a global community of robotics developers \cite{Macenski_2022}.

\section{Cloudification practice in ROS}
To enrich the ROS ecosystem, the open-source robotics foundation (OSRF) has been releasing docker container images  for different ROS distributions for years. This practice has gained popularity due to the inherent advantages of the container technology, such as encapsulation, environment consistency, and easier distribution of applications. However, the evolution towards cloudification, i.e., the integration of ROS with container orchestration platforms like \gls{k8s}, has been relatively slow. This transition would mean leveraging cloud-native tools and principles to provide simplicity, reliability, scalability, and observability to ROS-based applications, creating a truly cloud-native robotics platform. We explain the state of the art of the cloudification practice in the academic and industrial ROS community in the following subsections.

\subsection{Literature}
Integrating robotics applications into the cloud stems from the “cloud robotics” concept. In \cite{cloud_robotics}, cloud robotics was perceived as an evolutionary step after networked robotics. Initial architectures for cloud-robot interaction were proposed. The potential advantages of cloud integration were initially touched, and the challenges in terms of computation, communication, and optimization were analyzed, albeit at a preliminary level. 
The survey in \cite{survey_research_on_cloud_robotics} reviewed a series of early studies that proposed architectural design of cloud-based robotic systems for dedicated applications, such as robot grasping \cite{crowdsourcing} \cite{robot_grasping}, path planning \cite{robot_path_planning}, and SLAM \cite{roboearth}. These early-stage attempts often fall short in scalability and extensibility, making them hardly be used as a generic framework in cloud-robot practice. 

With the maturity of open-source projects in both robotic and cloud domain, ROS, container, and Kubernetes increasingly became the tools of choice for implementing cloud robotic applications. In \cite{internet_of_drone}, the authors proposed a cloud-based framework to provide cloud services to ROS-powered drone applications that are hosted on a Kubernetes cluster and exposed through URLs. In \cite{cloud_robot_app_platform}, the authors proposed a framework to enable locally deployed ROS nodes to exchange messages with ROS nodes in the Kubernetes cluster via rosbridge. In  \cite{k8s_based_edge_arch} \cite{comparison}, the authors developed an architecture to control the trajectory of ROS powered UAVs, and a model predictive controller is containerized and deployed on Kubernetes. Notably, all of these examples do not treat robots as part of the cluster, posing a significant challenge for robot-cloud communication.

Several approaches have been proposed to address the communication challenge. Technically, they can be divided into two categories: proxy-based \cite{rosremote} and VPN based approaches \cite{ros_through_vpn} \cite{5g_edge}, of which two most promising projects are FogROS2 \cite{fogros2} and its successor FogROS2-SGC \cite{fogros2sgc} proposed by UC Berkeley. 
Additionally, the authors of \cite{robotkube} propose to use Kubernetes to orchestrate ROS-based cooperative intelligent transport systems (C-ITS) where MQTT is adopted to facilitate communication among vehicles and cloud. The performance of these proprietary solutions under complicated network environment still needs further verification, apart from the deployment complexity.

The study in \cite{methodology} advocated for leveraging Kubernetes and Docker to modularize ROS applications and standardize the application deployment procedures.

\subsection{Industry}
From an industrial perspective, Canonical Ubuntu provided a series of blog posts to describe how to establish a ROS 2 talker and listener example in the Microk8s platform. This setup involves distributing multiple ROS 2 nodes across several machines in a \gls{lan} \cite{ros2_k8s_cononical}. 
Sony has also demonstrated the concept and architecture for integrating robotics into an edge cluster system . Preliminary implementation details have been revealed\cite{k8s_sony}, with both ROS 1 and ROS 2 docker images shown to be deployed to a Vanilla Kubernetes cluster.
RoboLaunch is a new startup which focuses on building a cloud robotics platform that offers end-to-end infrastructure and software stack to simplify development, simulation, and life-cycle management of robotics application. The cloud-based platform uses Kubernetes to orchestrate containerized ROS applications. A VPN is created among multiple clusters to enable cross-cluster communication and computation offloading. To date, this is still in the prototyping stage and not yet ready for production.

In conclusion, the exploration of incorporating ROS powered robotics applications to the Kubernetes platform has started in the industrial sector, but a mature, production-ready solution has yet to be seen. 
This paper aligns with the vision proposed in \cite{methodology} and aims to provide a framework, namely RoboKube, that simplifies the setup of a dedicated orchestration platform for ROS powered robotic applications for research and production purpose. RoboKube enables cross-network deployment in complex network environments, eliminating the fundamental barrier in cloud-robot communication, which opens the door to advanced robot-cloud and robot-network \cite{5g_qos} interactions.

\section{RoboKube: a platform perspective}
\label{sec:platform}
In this section, we delve into the specifics of how the RoboKube orchestration platform can be established. A key aspect in the discussion is around the networking and different container network backends utilized by RoboKube.

\subsection{Orchestration platform}
Orchestration platform plays an fundamental role in the cloud native domain, managing, scaling and ensuring the resilience of cloudified workloads. RoboKube does not rely on any specific orchestration distribution, i.e., any K8s compatible variants can be adopted. Considering simplicity of installation and compatibility to K8s ecosystem, \textbf{K3s} is one of the preferred solutions. 
K3s is a lightweight variation of K8s dedicated for resource-constrained edge computing and IoT user cases. K3s is packed in a single binary which largely reduces the dependencies and steps to deploy and manage a full-fledged K8s distribution. It retains the key capabilities essential for managing containerized applications and maintain the compatibility to K8s ecosystem to a great degree.

\subsection{Networking}
ROS 2 is built on top of \gls{dds}/\gls{rtps} which is an end-to-end middleware that provides features such as distributed discovery and control over different \gls{qos} options for the transportation. DDS/RTPS uses a brokerless pub-sub messaging system, and implements a reliable multicast over plain UDP sockets.
\begin{figure*}[tbp]
    \centering
    \includegraphics[width=0.8\linewidth]{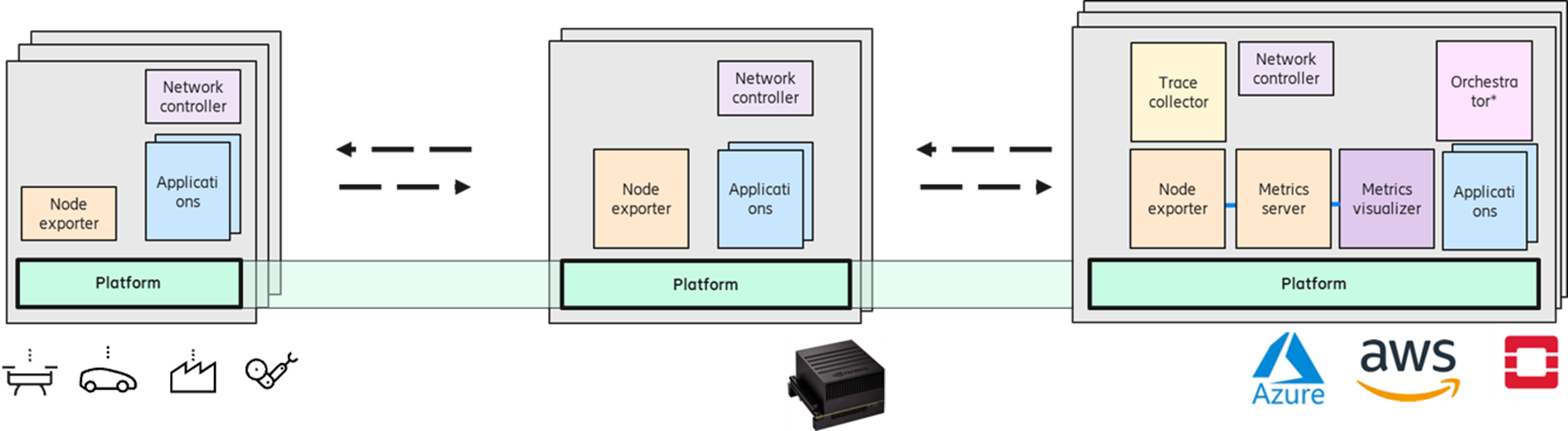}
    \caption{A common platform across the device-edge-cloud continuum is the cloud native approach to address the network heterogeneity issue.}
    \label{common-platform}
    \vspace{-2mm}
\end{figure*}
\subsubsection{Network backend alternatives}
RoboKube automates the network configuration and simplifies the network consideration of ROS 2 applications by introducing an overlay network across the device-edge-cloud continuum, as shown in \Cref{common-platform}. With properly selected network backends, through the \gls{cni} plugins, an overlay network is established for the orchestration platform which makes the complexities in the underlay network transparent to the applications and guarantees all the promising features of DDS, such as the pub-sub transport and node discovery, as long as multicast is supported by the network backend. Despite the diversity of different network backend implementations, only two are identified to support multicast. One of the preferred alternatives is \textbf{Kube-ovn}, which offers a tight integration between Open Virtual Network (OVN) and the container networking. It provides the capabilities of network virtualization to the K8s space, ensuring an extreme performance for the data plane.
\textbf{WeaveNet} is another alternative that enables UDP multicast, though the performance in terms of latency and bandwidth of both TCP and UDP traffic are worse than Kube-ovn, according to our benchmarking.

\subsubsection{Ingress and NodePort}
When a ROS 2 application needs to be exposed as a service to enable external accessibility through a specific group of ports of the host machine, RoboKube considers to use \textbf{Ingress} and \textbf{NodePort}. 

Both NordPort and Ingress offer an option to enable external traffic to access Kubernetes internal services. NordPort provides a nearly lossless solution that opens access ports across all cluster nodes, which is suitable for development and verification. Ingress provides a single point of entrance and a more flexible solution to configure routing rules for external traffic reaching out to internal ROS services, though with a minor loss in bandwidth. 

\subsubsection{Other network considerations}
In addition, the following network aspects need to be respected when setting up the RoboKube platform.
\begin{itemize}
    \item Due to the nature of Kubernetes, containers within a pod communicate over the loopback interface and share the same network stack. ROS 2 does not provide a method for managing ports used by RTPS, i.e., a ROS 2 container cannot change the standard RTPS discovery port of 7400, nor the default listener port. Consequently, port usage is not coordinated when multiple ROS 2 containers are running in the same pod, therefore, these containers are not able to communicate . As a result, in practice, each pod should only run a single ROS 2 container inside to avoid the issue. Within a container, multiple ROS 2 nodes can be deployed.
    
    \item In Kubernetes, Services  is a common method to expose applications in a pod. However, K8s Service need to perform port or address translation which interferes with ROS 2 communications, therefore cannot be used for ROS 2 network traffic.
    
    \item The multicast message is sent to every K8s node no matter if there are ROS 2 nodes or not. This behavior can potentially incur an overloading problem. In the RoboKube platform setup, we propose to enable Internet Group Management Protocol (IGMP) snooping to avoid unnecessary Mcast packets.
    
    \item In the RoboKube orchestration platform, the maximum transmission unit (MTU) sizes used in the overlay and underlay network need to be coordinated. If not properly set, it can result in packets being dropped and ignored. In practice, the size of Pod MTU can be 100 bytes less than the physical interface MTU size.
\end{itemize}

\section{RoboKube: an application perspective}
\label{sec:application}
As described in the sections before, RoboKube establishes the cloud native platform to be able to deploy cloudified application and simplify the process. However, the questions of packaging ROS-based applications remain open. This section tries to elaborate the identified considerations and best practices.

\subsection{Containerization of ROS 2 nodes}
There exist a series of best practices for building container images. For the scope of this paper, we primarily focus on two requirements:  1) usability, which is to devise a simplified methodology that empowers ROS developers to construct images, and 2) scalability, which emphasizes to provide low-footprint images to increase resource efficiency and enable large scale deployment.

One recommended option to optimize the ROS application image is to use a third-party toolkit namely \textbf{DockerSlim}.
DockerSlim can dynamically probe the applications running inside a container during runtime, recording all necessary libraries and dependencies while removing those unused components, which results in up to 30x smaller container image. Meanwhile, the whole process is highly automated, if the applications to be executed are properly specified during container launch time. 

\subsection{Deployment}
The simplest way to run a containerized ROS node is to directly launch the docker image. This straightforward method is ideal for running a single node or a group of mutually interactive nodes that are launched in the same container and no connections or dependencies to external nodes are needed. However, this approach lacks the orchestration capabilities. In practice, docker based deployment approach is more for the test phase and function verification but not for production deployment of ROS 2 applications.

As a package manager for Kubernetes, \textbf{Helm} is the proposed way to deploy containerized ROS 2 applications on RoboKube platform. Helm simplifies the deployment process by providing templated applications. This feature enables multiple deployments and streamlines the management and versioning of applications, which opens the opportunity to realize live migration / computation offloading of ROS 2 components across nodes.

\subsection{ Distribution and clustering of ROS nodes}
Decisions on distribution and clustering of ROS nodes are largely dependent on the specific application and can mutually influence each other. Below we preliminarily summarize several aspects for application developers to consider. 
\begin{itemize}
    \item \textbf{Hardware affinity}: it refers to those that enforce a ROS node to be assigned to a specific Kubernetes node due to the need to access a specific hardware, service, and storage, etc., or due to the consideration of data and privacy preservation.
    \item \textbf{Performance metrics}: it can include application layer metrics such as mean average precision (mAP) for objection detection, absolute position error for SLAM, or end to end execution time, etc. These metrics can further be impacted by system metrics like network latency and resource utilization. To cater to the performance requirement, sometimes we need to perform profiling of different distribution and clustering solutions to understand its implications on performance metrics and identify the constraints.
    \item \textbf{Offloading and migration capability}: Taking computation offloading into account, how to distribute and cluster ROS nodes can vary a lot in the static and the dynamic offloading modes. In the static mode, the topology of ROS nodes is rather stable once the application is scheduled. In the dynamic mode, a friction of ROS nodes may be migratable thus need to be split into a single module. However, a fine-grained splitting of node implies high flexibility in offloading but may introduce extra complexity and overhead due to frequent container migration. Therefore, a tradeoff between offloading flexibility and other factors has to be maintained.
\end{itemize}

\section{Case study: RoboKube-powered teleoperation testbed}
\label{sec:case}
\begin{figure}[tbp]
    \centering
    \includegraphics[width=0.7\linewidth]{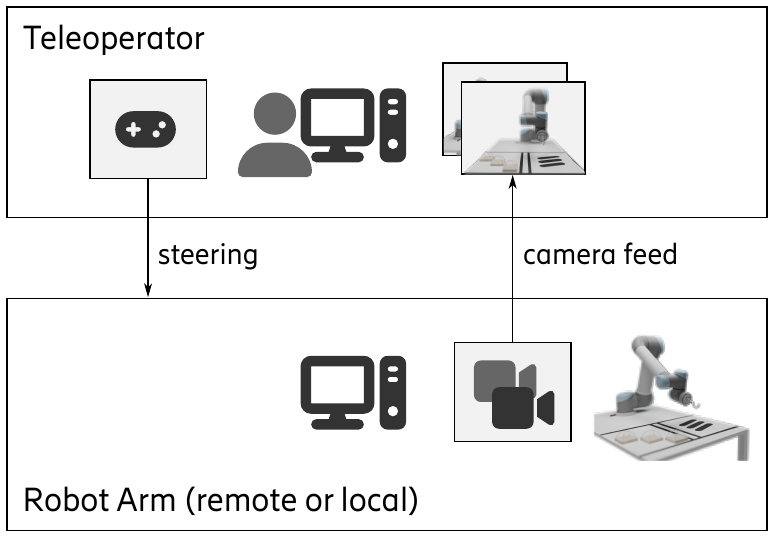}
    \caption{The high-level architecture of the teleoperation testbed.}
    \label{teleop-arch}
\end{figure}

 In this section, we demonstrate a case study of a teleoperation testbed that is set up using the proposed RoboKube framework. In the testbed, a Universal Robots UR5 robot arm can be teleoperated at a distance using a joystick. The architecture is illustrated in \Cref{teleop-arch}.

\subsection{Application overview}
\begin{figure}[tbp]
    \centering
    \includegraphics[width=\linewidth]{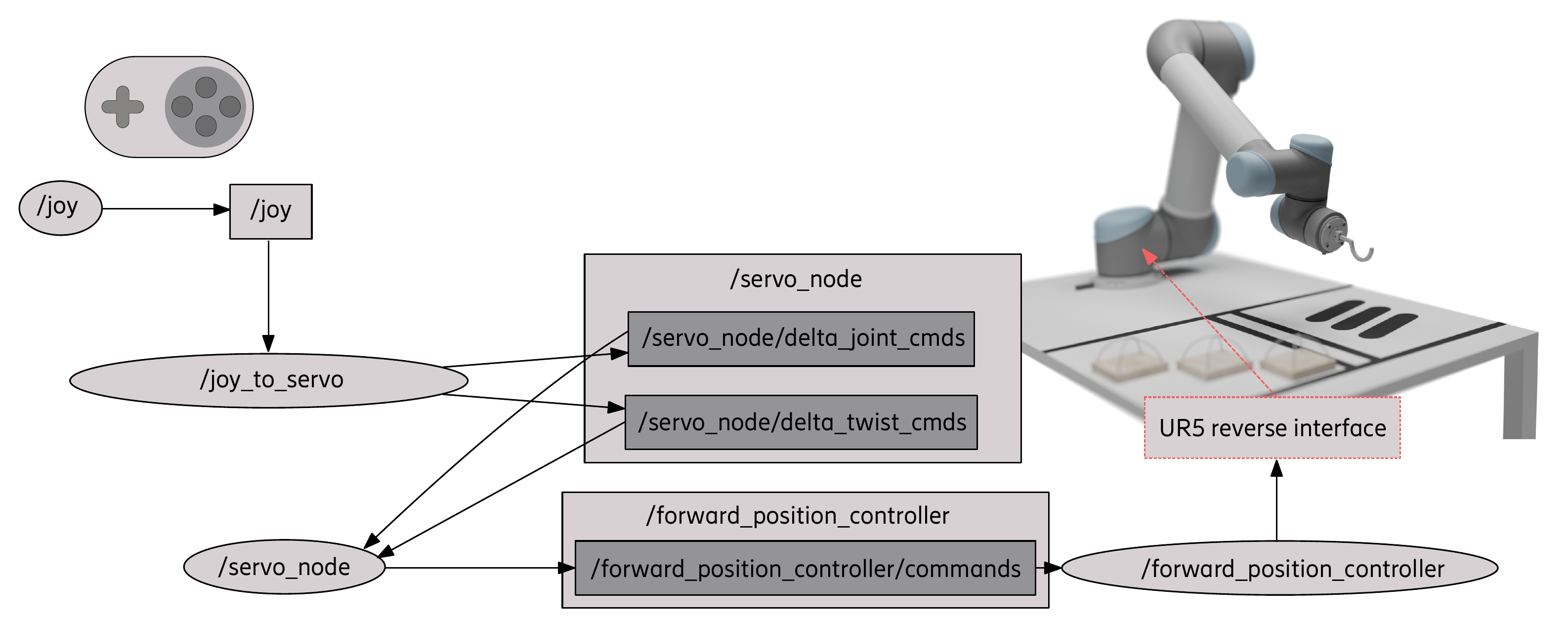}
    \caption{The ROS 2 architecture, where ellipses denote ROS nodes and rectangles denote ROS topics. Each ROS node except joy can be distributed and run on an arbitrary Kubernetes node in the cluster.}
    \label{teleop-ros2-arch}
    \vspace{-5mm}
\end{figure}

The high-level structure of the application can be seen in \Cref{teleop-ros2-arch}, which depicts the ROS nodes and their communication through ROS topics. In each end of the graph, we have hardware connections: the joystick at the top of the graph, and the Universal Robots UR5 manipulator at the bottom.
Continuing down the graph, we find servo\_node, which is a node that relies on MoveIt to translate either joint or end-effector velocities into a corresponding desired joint position, which ultimately is handled by the forward\_position\_controller node that publishes to a topic handled by the UR5 reverse interface.

In this testbed, each ROS node can be distributed onto any Kubernetes node in the cluster apart from one exception: the joy node. This is due to that the joy node relies on USB hardware, namely, the joystick itself. While the UR5 itself also belongs to the hardware category, it is configured as a network device. Hence, the UR5 reverse interface can be placed anywhere in the cluster, just like any other ROS node.


\subsection{Containerization and distribution of ROS nodes}
During the containerization process, the teleoperation application is split into two containers, the joystick container that is dedicated to running the joy node and the UR5 driver container that executes the servo\_node and the forward\_position\_controller node. The splitting is due to that the joy node has to be deployed to the machine with physical joystick hardware while the other nodes can be distributed across the cluster as per the orchestrator’s runtime decision.

Containerization of the application is done by a multi-stage build. Taking the joy node container for example, a two-stage build process is applied. In the first stage, ROS dependencies are installed, and the joy node application is built from source. After that, all the libraries that are required by the joy node as well as the associated libraries (i.e., libjoy\_to\_servo) are manually extracted with the “ldd” utility. In the second stage, a minimal ROS docker image, e.g., the ROS base image, is utilized as the base image for the application. The application executable and associated libraries, the previously extracted dependency libraries as well as the ROS stack libraries are copied from stage 1 to this stage. In this way, the joy node gets a qualified running environment with all necessary dependencies while keeping the footprint of the container image relatively small.
The achieved image is further shrunken with the assistance of DockerSlim, which only keeps the joy node binary and the dependencies that are used during the node execution. All other components including system utilities, shell, unused system libraries, and other redundant have been removed from the image. In the end, the image size is reduced by 82\% from 486 MB to 83 MB, lowering the footprint to a great extent. The same approach can be applied to the UR5 driver container, of which the image merely accounts for ~300 MB compared to the original size of 2.6 GB.

\subsection{Deployment}
As mentioned in \Cref{sec:platform}, a Kubernetes cluster is created for the teleoperation testbed, with two Kubernetes nodes distributed in two different subnets for demonstration purpose. Deployment to the cluster is handled by Helm Chart which enables rolling update and rollback of the application. Additionally, some details are revealed as follows.
\subsubsection{Device plugin}
with the use of device plugins, a USB joystick controller connected to a node can be abstracted as a resource. Consequently, this resource can be allocated to pods using the "requests" and "limits" specifications, as shown below.

\begin{lstlisting}
resources:
    limits:
      squat.ai/joystick: 1
\end{lstlisting}
In this way, the joy node container can only be deployed to a node with a joystick connected, making the scheduling hardware dependent.

\subsubsection{Ingress}
As the UR5 driver launches, it will listen on port 50001 and 50002 , waiting for the robot to get connected. Due to the default limit in Kubernetes that NodePort can only be assigned within the range 30000-32767, Ingress (e.g., Traefik Ingress) becomes a more suited solution to open the needed ports. 

The teleoperation testbed can be cloudified and deployed to the RoboKube orchestration platform, which is able to function in a WAN network. The deployment and upgrade can be achieved in one step with Helm Chart while all components can be freely migrated in any K8s nodes except the joystick node. This case study demonstrate how the proposed RoboKube framework can be utilized in research and production environment.

\section{Concluding remarks}
\label{sec:conclusions}
As cloud technologies mature and expand from the cloud industry into the IoT and CPS domain, it is observed that the integration of robotics applications into cloud has started in the research community, but it is far from attaining widespread use in industrial practice. This paper bring outs RoboKube, a (work-in-progress) framework that intends to bridge the gap and facilitate the integration between cloud technologies and the ROS ecosystem. It provides comprehensive solution to create a Kubernetes based platform for cloudified ROS applications, with an emphasis on the network setup to enable deployment in heterogeneous network environments that can include wireless and cellular networks. It also aims to give guidance on the best practices of containerization approach and deployment solutions, as well as the factors to be considered when distributing and clustering ROS nodes. The paper aims to make the cloudification of ROS-powered applications an achievable reality so as to accelerate the cloud native evolution in robotics.

\bibliographystyle{IEEEtran}
\bibliography{main.bib}
\end{document}